# Automatic pre-grasps generation for unknown 3D objects

IA Sainul
Advanced Technology Development Centre, IIT Kharagpur,
Kharagpur -721302, India
sainul@iitkgp.ac.in

Sankha Deb
Mechanical Engineering Department, IIT Kharagpur,
Kharagpur -721302, India
sankha.deb@mech.iitkgp.ac.in

AK Deb
Electrical Engineering Department, IIT Kharagpur,
Kharagpur -721302, India
alokkanti@ee.iitkgp.ac.in

*Abstract*—In this paper, the problem of automating the pre-grasps generation for novel 3d objects has been discussed. The objects represented as a cloud of 3D points are split into parts and organized in a tree structure, where parts are approximated by simple box primitives. Applying grasping only on the individual object parts may miss a good grasp which involves a combination of parts. The problem has been addressed by traversing the decomposition tree and checking each node of the tree for possible pre-grasps against a set of conditions. Further, a face mask has been introduced to encode the free and blocked faces of the box primitives. Pre-grasps are generated only for the free faces. Finally, the proposed method implemented on a set twenty-four household objects and toys, where a grasp planner based on object slicing method has been used to compute the contact-level grasp plan.

## I. INTRODUCTION

Automatic grasping is a challenging problem in robotics particularly, for service robots which encounter with commonly known and unknown objects in house/office environments. Challenges like, mathematical complexity, varying degrees of uncertainty in the object perception, make sure that the current advancement in the field nowhere near to the human counterpart and still remains an active area of research. Humans tend to place the hand and its fingers to an appropriate prehensile posture chosen for particular object geometry and then close the fingers to grasp the object. A similar approach is used in robotic grasping, where pre-grasps are generated before computing the actual contact-level grasp plan. The pre-grasp basically consists of initial gripper/hand position, approach direction and its finger configuration depending upon the object geometry. The contact-level grasp planning computes the contacts between the gripper/hand and the object which ensure stable grasp. The initial grasp configuration or pre-grasp depends on the object shape, rather than identification of the object. Object part based grasp selection had been used in the past where the object was decomposed manually and the parts were approximated using primitive shapes e.g. planes, boxes, cones, spheres or cylinders [1]. Goldfeder et al. [2] used Super Quadrics (SQs) as generic shape primitives to automate the process and organized the object parts in a decomposition tree. More, recent work [3] had argued that the successful grasp selection depends upon the geometry rather than the object identification and put more emphasis on the efficiency over the accuracy in the shape approximation. One problem associated with the grasp selection for individual part is that the part may not be fully accessible to the gripper/hand due to occlusion by the neighbouring parts. Another problem of applying the part based method only on the individual object parts is that it may miss a good grasp which involves a combination of parts. While, the first problem has not been covered in the literature, some attempts were made to address the second problem. The objective of this work is focused on addressing the two research gaps and automating the pre-grasps generation for novel 3d objects.

In the following, some of the previous works related to the proposed approach are discussed. In the medical field, Napier [4] classified all the prehensile postures used by human hands for grasping different object geometries into grasp taxonomies. Cutkosky and Wright [5] attempted to classify hand postures and find grasp taxonomies needed for robot gripper in a manufacturing cell. The pre-grasp generation requires the gripper/hand to be correctly positioned and oriented relative to the object for fingers to reach the object. Besides the complexity in the object geometry, the hand internal degrees of freedom and those in the wrist of the hand create a huge grasp search space. The internal degrees of freedom (DOFs) in the hand along with 6 DOFs in the robot arm make the grasp space too large to be exhaustively searched. Many approaches are there in the literature [6, 7] to find good wrist position and orientation in this huge search space. Predefined prototypes of grasp have been used to reduce the search space [8]. Hester et al. [9] reduced the size of the grasp search space by assuming fixed wrist position. A single view 3D point cloud data taken using a depth camera was approximated by simple box primitives [3]. Then heuristics used to select graspable box faces and finally, an off-line trained neural network gives the best grasp hypothesis. Li et al. [10] employed deterministic sampling on the spherical surface constructed around the object to find the initial hand position and approach direction. Shape diameter function (SDF) was used by Vahrenkamp et al. [11] to segment objects into parts and then principal component analysis (PCA) was applied on the parts to align the hand with the corresponding object part. The main drawback of the method is that it can only be applied on object polygon mesh.

In this paper, the problem of automating the pre-grasps generation for novel 3d objects has been discussed. First, the objects represented as a cloud of 3D points are split into parts and organized in a tree structure where parts are approximated by simple box primitives. Then, Principal Component Analysis (PCA) is used to classify the parts and assigned appropriate type of grasps. Next, feasible surface patches around the object parts, where the gripper can reach, are sampled depending upon the type of grasps and the final pool of pre-grasps is generated. Finally, an object slicing based grasp planner [12] is applied to find contact-level grasp plan for illustrating the complete results.

**Contributions:** The important contributions reported in this paper are as follows. The proposed method can handle objects with complex shapes and automatically generates pre-grasps; it does not require expensive pre-processing like object surface reconstruction and works on point clouds taken using depth sensors; it generates only feasible pre-grasps where the gripper can reach for both adaptive/enveloping and fingertip types of grasps.

The rest of the paper is organized as follows. Section II gives an overview of the three finger gripper. Section III presents the pre-grasp generation. Section IV presents the implementation results. Finally, Section V gives the conclusions.

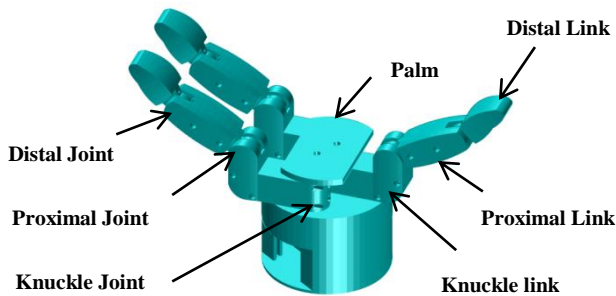

Fig. 1. Model of the robotic gripper

## II. ROBOT HANDS

In this section, an overview of a three finger underactuated robotic hand and its actuation mechanism is given. Overall hand mechanism and actuation mechanism of the fingers are discussed in [13]. A model of the robotic hand is shown in figure 1. The hand consists of three identical fingers each having three links namely knuckle, middle, and distal phalanx. The thumb finger has two joints and its base is fixed on the palm. One more joint is incorporated at the base of each of the other two fingers which enables it to spread sideways synchronously around an axis perpendicular to the palm surface. The two joints of the fixed finger and three joints of each of the other two fingers are actuated by a total of four DC motors. The synchronous spreading motion of the two fingers is achieved by placing one servo motor and a gear system inside the palm. A tendon-pulley system is used in each finger to achieve the flexion and extension motion of the middle and distal phalanx of the finger.

## III. PRE-GRASP GENERATION

This work is mainly focused on automating robotic grasping of 3D objects having arbitrary shapes and sizes. The proposed grasp planner assumed that the complete information about the object geometry is available. First, the object is decomposed using simple box primitives and a pool of pre-grasps is generated for the object parts. Where, pre-grasp consists of initial hand position, approach direction and finger configuration of the hand. Then, an object slicing based method [12] is applied to quickly find the contact points on the object for evaluating the quality of the grasps. Finally, all the contacts and finger joint displacements are computed by closing fingers until contacts are found where the actual mesh model of the hand is considered.

### A. Object Decomposition

The data captured using a dense stereo camera or a depth sensor is a cloud of 3D points, which is further processed to reconstruct the object and stored as a polygonal mesh. It is very difficult to come up with a grasp strategy from this kind of low-level representation of the object. A 3D object can be represented using a set of shape primitives (e.g. planes, boxes, spheres or cylinders). A more generic approach is to use Super Quadrics (SQs) as shape primitives which offer a large variety of different shapes. This kind of high-level representation is very useful for operations like grasping where appropriate grasp strategies can be applied to different shapes. Complex shape primitives such as super quadrics give good approximation of the object but difficult to process. For grasping, efficiency is more important than the approximation accuracy. So, simple boxes are used for approximating the objects as shape primitives and then types of grasp are selected as per the criterion described in the following sub-section B. The object decomposition is done by using a fit and split algorithm based on Minimum Volume Bounding Box (MVBB) [14]. The output of the object decomposition algorithm is a set of Oriented Bounding Boxes (OBBs) organized in tree structure. The algorithm starts with tightly fitting the data points by a bounding box having minimum volume. Then, it iteratively splits the box and its data points such that new point sets yield better box approximations of the shape. The splitting step is about finding a plane which results a good split of the parent box. A good split is the one which minimizes the summed volume of the two resulting child boxes. Such method demands extensive search and comparison of a lot of planes with different position and orientation to find a good split plane. Therefore, only the planes parallel to the parent box are considered for the potential partitioning. The splitting is carried out iteratively until a box is not dividable or reaches minimum data points as shown in figure 2.

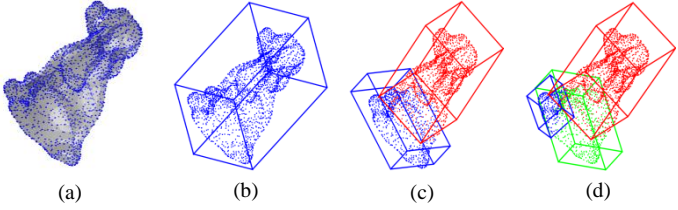

Fig. 2. (a) Point cloud (b) Bounding box, (c) Two child boxes resulted from root box in the first iteration, (d) Final decomposed object approximated by three boxes.

### B. Grasp Type Selection

The idea of dividing an object into a decomposition tree is to capture local shape descriptions and apply an appropriate type of grasp to the respective part of the object. Moreover, it enables the use of efficient global shape descriptor like Principal Component Analysis (PCA). On the contrary, it is not possible if the object is considered as a whole. In this subsection, the object parts are classified into number object categories using PCA. Then appropriate grasp strategies are assigned to each part of the object. Principal Component Analysis (PCA) on the data points of an object gives information about the object distributions. Let the eigenvalues of the PCA be denoted as $\lambda_1$, $\lambda_2$ and $\lambda_3$ where, $\lambda_1 \geq \lambda_2 \geq \lambda_3$. Then the object is classified into number of categories, for example, principal component of one-dimensional object is significantly larger than the other components i.e. $\lambda_1 \gg \lambda_2 \geq \lambda_3$, for two-dimensional objects $\lambda_1 \cong \lambda_2 \gg \lambda_3$, and for three-dimensional objects $\lambda_1 \cong \lambda_2 \cong \lambda_3$. Further, actual dimensions along the principal components are used to differentiate larger and smaller three-dimensional objects.

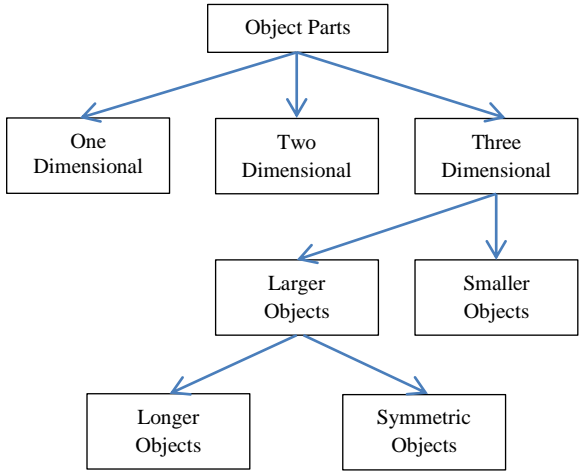

Fig. 3. Object Classification based on object shape information using principal component analysis (PCA)

Different finger configurations enable the hand to handle a wide range of possible form/force closure grasps. Depending upon the size and shape of the objects shown in Figure 3, a total of four types of grasps has been recognized for this work e.g., cylindrical grasp, spherical grasp, three-fingered tip grasp, and two-fingered tip grasp.

**Cylindrical grasp:** A three-dimensional long object can be grasped using this configuration. In this configuration, the palm provides support to the object and all the fingers wrap around the object from two opposite sides as shown in figure 4(a). Sideways spreading movement is not required for this type of grasp. Other similar objects can be grasped using this method.

**Spherical grasp:** Generally, three-dimensional symmetric objects can be grasped using this configuration. Similar to cylindrical grasp, the palm provides support to the object. In this grasp, the two fingers spread sideways and place all the fingers in symmetric position around the object and wrap around the object as shown in figure 4(b).

**Three-fingered tip grasp:** Two-dimensional objects can be grasped using the fingertips of all the three fingers from two opposite sides of the object as shown in figure 4(c).

**Two fingered tip grasp:** Small three dimensional objects can be grasped using the fingertips of the two spreading fingers. This type of grasp requires only two fingers. The two fingers spread sideways up to $90^0$ and place itself on the opposite side of the object as shown in figure 4(d).

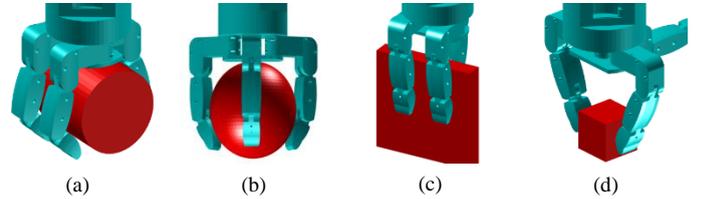

Fig. 4. (a) Cylindrical object grasp, (b) Spherical object grasp, (c) Two-dimensional flat object grasp, (d) Smaller object grasp using fingertips.

### C. Reachability, Blocking and Face Mask

Let, an object is a set of 3D points $P \subset \mathbb{R}^3$, then the decomposition technique partitions an object into a set of object parts $P = \{P_1, .. P_n\}$ and are enclosed by a set boxes $B = \{B_1, ..., B_n\}$, where, $n$ is the number of partitions. The type of grasp for each of these boxes is selected from the four types of grasp. Although each type of grasp puts some constraints on the gripper in terms of finger configuration and alignment with the object, the gripper still can approach and align itself in a large number of ways. It is important to find a way to reduce the number. This is where the reachability of the gripper comes into consideration. A gripper can reach a box from its six rectangular faces. Now, the two-child boxes of a parent box share common faces and some faces may be occluded by neighbouring boxes in the decomposition tree, so all the faces cannot be reached by a gripper. Further, such faces block fingers when the griper tries to reach from its adjacent free face. As shown in figure 5(a, b) where each face of a box has four adjacent faces and can be in two states either free or blocked denoted by white and black colours. The states of the faces are coded in face-mask matrix (0 for free and 1 for blocked), where each row denotes a face and its four adjacent faces as shown in figure 5(c).

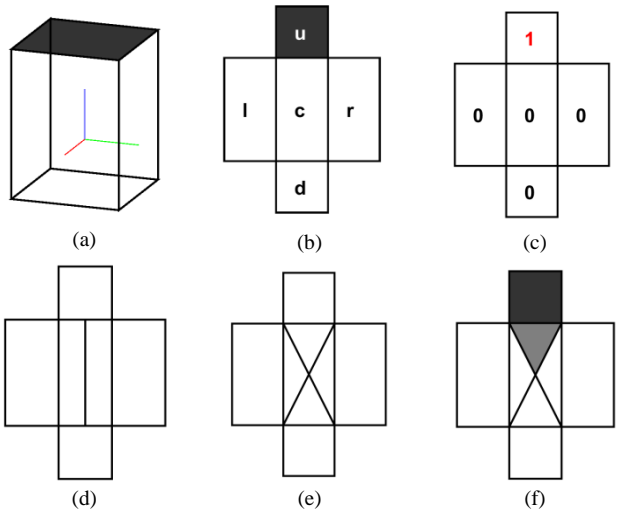

Fig. 5. (a) Bounding box with a blocked face coloured black, (b) Centre face having four adjacent faces left, down, right, and up denoted by (l, d, r, and u), respectively, (c) Face-mask matrix coded with 0 and 1 for free and blocked, respectively, (d) & (e) Sub-face schemes for cylindrical and spherical types of grasps, (f) An example of how a blocked face eventually blocks an adjacent sub-face.

At the time of pre-grasps generation, the gripper is aligned only with the free faces. In addition, a blocked adjacent face eventually reduces the free face area e.g., a gripper cannot orient its fingers towards the up face (u) when it is at centre face (c) near to the up face (u) as shown in figure 5(b). So, to effectively avoid such situations the face is divided into sub-faces and associated with adjacent faces. Only those sub-faces having free adjacent faces are considered for pre-grasps generation. The division depends upon the types of grasp as shown in figure 5 (d, e).

*D. Generation of Pre-Grasp Pool*

In this sub-section, a pool of pre-grasps is generated by sampling enclosing surface areas around the object. A pre-grasp consists of an initial gripper position, orientation and its finger configuration. The orientation gives approach direction of the gripper towards the object. The sampling surface is predefined for each type of grasps. First, a surface is created which fully encloses the bounding box of the object part. Then all the sub-faces having free adjacent faces are found out by using the face-mask and projected onto the created surface. The projected areas are sampled to get the gripper initial position and a vector towards the centre of the box gives the gripper approach direction. For the spherical grasp, the sub-faces are projected onto the surface of a sphere enclosing the object part and the projected sub-faces are sampled at a fixed interval, where the hand approaches along the radial vector as shown in figure 6(a, b). The cylindrical surface enclosing the object part is used to project the sub-faces for the cylindrical type of grasps as shown figure 6(c, d), where the hand moves along the radial-vector for the circular surface and axial-vector for the flat surface of the enclosing cylinder. An enclosing circle is sampled for the three-fingered tip type of grasp and the hand is oriented radially as shown in figure 6(e, f), Similar to spherical grasp, a spherical surface is used for the two-fingered tip type of grasp.

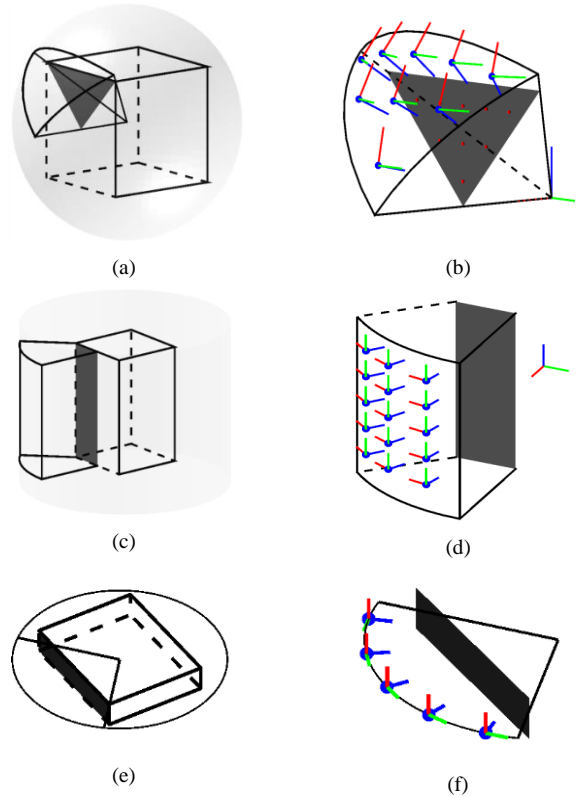

Fig. 6. Different sampling strategies for the generation of the pre-grasp pool.

The disadvantage of applying the said method only on the individual object parts is that it may miss a good grasp which involves a combination of parts. So to avoid such a scenario, the process is iterated over the object decomposition tree starting with the parts at the root and go downwards. At each node, two conditions are checked to decide whether further steps will be done or not. Firstly, if it has no child nodes or one of the children is of the type of small object then only further steps are carried out for the current node to generate pre-grasps. Secondly, actual dimensions are checked against a threshold value to decide whether the gripper can hold that particular part or not. For example, there might be a situation where a big part cannot be fitted in a gripper but its child parts can be separately fitted. The second condition successfully prevents big part but passes the children for the further steps of the pre-grasp generation. All the samples generated from the decomposition tree are stacked to form the pool of pre-grasps for an object.

## IV. GRASP PLANNER

An object slicing based method [12] is used to quickly find grasp points on the object. The grasp planner takes the pool of pre-grasps as input and computes contacts between the fingers and the objects. Once, all contacts between the fingers and the object are determined, the grasps are evaluated for stability using grasp quality measure [15] and rank all the grasps in the pre-grasp pool and then the best grasp is chosen.

## V. RESULTS

The proposed pre-grasps algorithm is implemented on objects that are taken from the KIT grasp benchmarking database [16]. The KIT database was made of scanning real household objects and toys using a 3D depth sensor and a stereo camera. Here, only the results of four objects for various types of grasps with diverse shapes and sizes from the total of 24 objects are given.

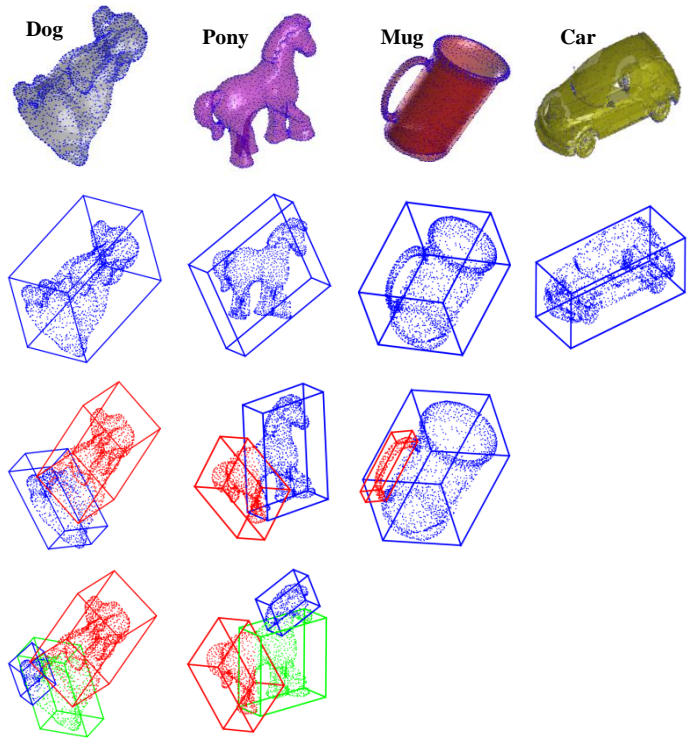

Fig. 7. Object decomposition results at different splitting stages.

Figure 7 shows the object decomposition results at different splitting stages, where the parent box volume to the child boxes summed volume ratio of 0.9 and minimum data of 500 points in a box are used for termination condition of the splitting. The decomposition tree has 3, 3, 2, and 1 number of parts for the dog, pony, mug, and toy car, respectively. Figure 8 shows an example of the tree traversal for the pre-grasp generation which starts at the root and goes downwards. The final pre-grasp pool is shown in figure 9 where pre-grasps for each part in the decomposition tree are assembled to form the final pool. In case of small object (e.g., toy car as shown figure 10), no object decomposition is needed and pre-grasp generation is applied directly on the object. The final results of gripper grasping the object using the best grasp are shown in figure 10.

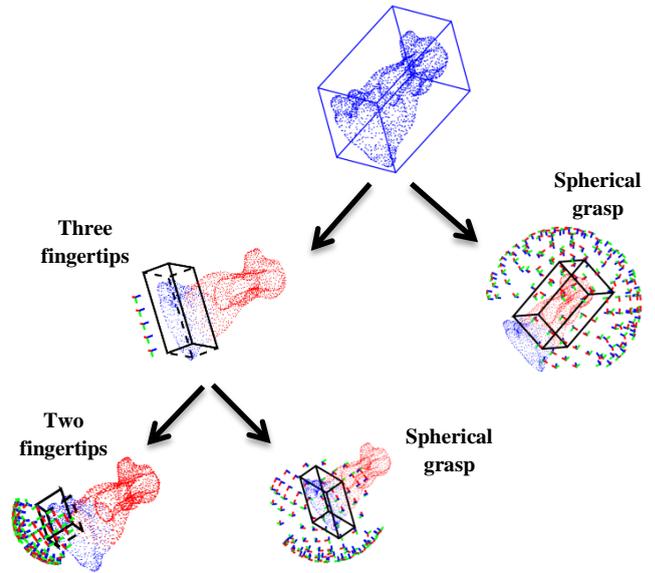

Fig. 8. Type of grasp selection and pre-grasps generation at different stages of the decomposition tree

In the example of a dog model, three types of grasps can be applied as shown in figure 9, but only spherical type of grasp has been automatically selected by the grasp planner. It is because of the three fingers involving in the spherical grasp make more number of contacts with the object than the other types of grasps and more number of contacts make the grasp more stable. Similarly, three fingertip grasp is chosen over two fingertip grasp for the pony and Cylindrical grasp is chosen over two fingertip grasp for the mug.

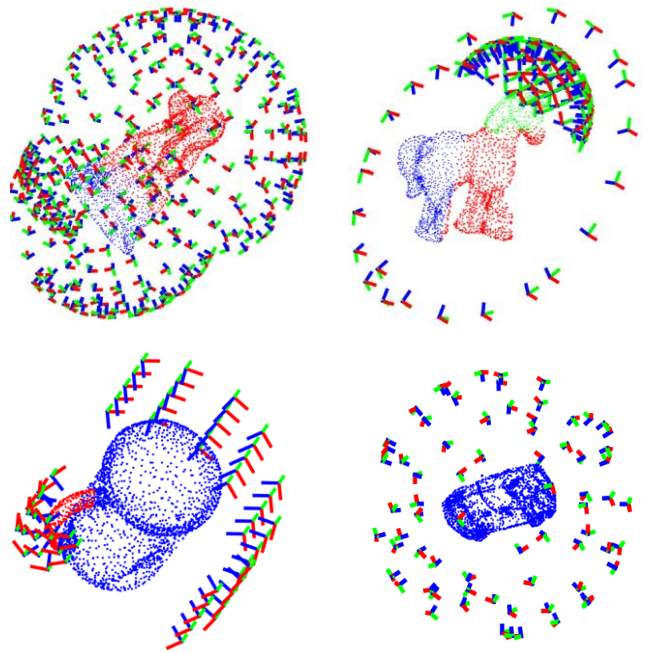

Fig. 9. The final pools of pre-grasps for the four examples

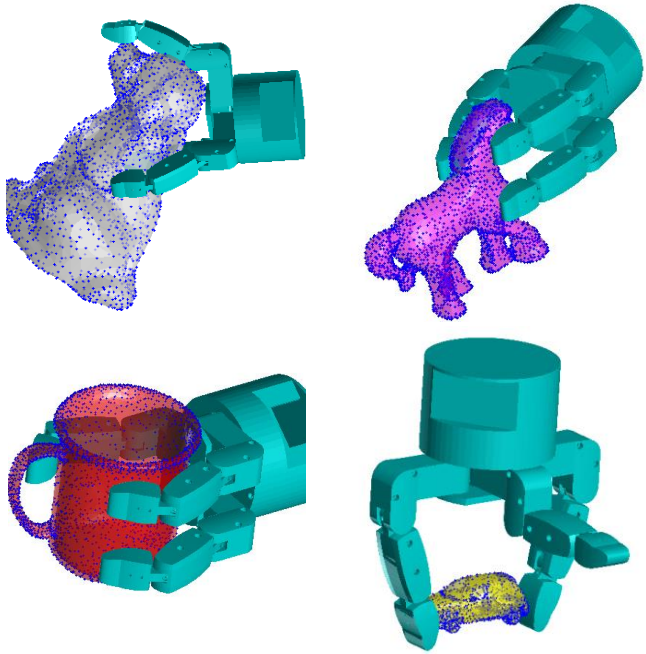

Fig. 10. Final results of grasping of various objects using the best grasp strategy computed using the contact-level grasp planner

## VI. CONCLUSION

In this paper, a framework for generating pre-grasps has been developed and implemented using a three finger underactuated robot gripper. Common household objects and toys have been used to validate the proposed framework and show its relevance in the service robotics field. Further, a grasp planner based on object slicing has been used to compute the final contact-level grasp plan. In an attempt to accomplish automatic grasping, the following are some important findings of the work: it can handle objects with complex shapes and sizes; it can be applied on point clouds taken using depth sensor; it takes into account self-occlusion of the object neighbouring parts and generates only feasible pre-grasps where the gripper can reach for both adaptive/enveloping and fingertip types of grasps.